# A Novel Directional Weighted Minimum Deviation (DWMD) Based Filter for Removal of Random Valued Impulse Noise

**Prof. J. K. Mondal**
**Department of Computer Science and Engineering**
**Kalyani University, Kalyani**
**West Bengal, India**
e-mail: jkm.cse@gmail.com
Phone: +91-33-25809617

Somnath Mukhopadhyay
M.Tech (CSE), 2nd Year
Department of Computer Science and Engineering
Kalyani University, Kalyani
e-mail: som.cse@live.com

*Abstract---The most median-based de noising methods works fine for restoring the images corrupted by Random Valued Impulse Noise with low noise level but very poor with highly corrupted images. In this paper a directional weighted minimum deviation (DWMD) based filter has been proposed for removal of high random valued impulse noise (RVIN). The proposed approach based on Standard Deviation (SD) works in two phases. The first phase detects the contaminated pixels by differencing between the test pixel and its neighbor pixels aligned with four main directions. The second phase filters only those pixels keeping others intact. The filtering scheme is based on minimum standard deviation of the four directional pixels. Extensive simulations show that the proposed filter not only provide better performance of de noising RVIN but can preserve more details features even thin lines or dots. This technique shows better performance in terms of PSNR, Image Fidelity and Computational Cost compared to the existing filters.*

**Key Words—Directional Weighted Minimum Deviation Filter (DWMDF), Image de noising, Random Valued Impulse Noise (RVIN), Peak Signal to Noise Ratio (PSNR), Image Fidelity (IF).**

## I: INTRODUCTION:

The corruption of images by impulse noise is frequent problem during image acquisition and transmission. Attenuation of noise and preservation of fine details are usually two contradictory aspects of image processing. The nonlinear characteristics of the impulse noise lead to poor performance of standard linear filters. So a series of nonlinear filter has been introduced to counter such noise. One of the most popular nonlinear filters is median filter[1,2,3,5,6,7,8,11]. The statistical and robustness property of the median filter with low computational complexity has made it suitable for de noising. But the main drawback of the median filter is that it works well for salt and pepper noise but not for images corrupted highly with RVIN and another thing is it also modifies the noise free pixels, thus it removes the fine details and to some extend it lost the faithfulness (Fidelity) of the original image. Impulse noise has the characteristic of corrupting only a certain percentage of image pixels leaving others unchanged. Further the gray values of the contaminated pixels are drastically different from the gray values of their neighboring pixels. The primary objective in RVIN removal is to de noising as well as to preserve the fidelity of the image[4,9,10].

In order to improve the median filter, many filters with an impulse detector has are proposed, such as signal-dependent rank order mean (SD-ROM) filter [7], adaptive center-weighted median (ACWM) filter [3], multi state median (MSM) filter [6] and the pixel-wise MAD (PWMAD) filter[5]. These filters usually perform well but the noise level is higher than 30%, they lost the faithfulness of the original image in the reconstructed image.

In a recently proposed paper, directional weighted median (DWM) filter [7] uses 8 to 10 iterations to noisy image to make it noise free. This filter uses new impulse detector, which is based on the differences between the test pixel and its neighbor pixels aligned with four main directions which uses a total of 16 neighbor pixels. And in a more recent scheme a second order difference based impulse detection method is used which has little less computational cost as because it uses only 8 neighboring pixels but it has a drawback that it does not detect a grater or equal number of noisy pixel with highly corrupted image. Extensive experiment on various images shows that DWM impulse detection method can perform better in presence of high percentage of impulse on image. But the computational cost(overhead) of impulse filtering in case of DWM filter makes it little inefficient.

So in this paper the DWM impulse detection method has been followed and a new filtering scheme based on Minimum Standard deviation, which uses a single iteration and a uniform threshold value for impulse detection for all types of images. Proposed scheme for filtering RVIN in images gives good Fidelity and a satisfactory PSNR value compared to existing schemes even when the images being highly RVIN corrupted.

Rest of the paper is organized as follows. The impulse detector is illustrated in section II. The proposed filtering scheme is formulated in section III. Experiment results and discussions are demonstrated in Section IV. Conclusion is given in Section V.

## II: DWMD Impulse Detector:

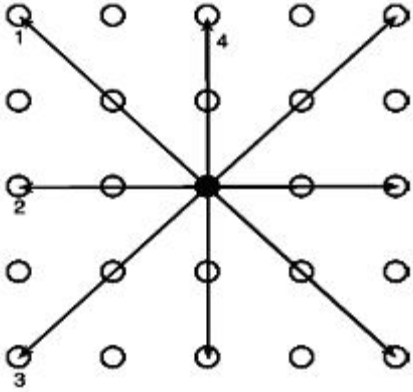

Fig. 1. Four directions for impulse detection.

The DWM impulse detector assumes that is a noise free image consists of locally smoothly varying areas separated by edges. In this scheme we have focused on the edges aligned with four main directions shown in Fig. 1.

Let $S_k$ (k=1 to 4) denote a set of coordinates aligned with the $k^{th}$ direction centered at (0, 0), i.e.,

$S_1$= {(-2,-2), (-1,-1), (0, 0), (1, 1), (2, 2)}
$S_2$= {(0,-2), (0,-1), (0, 0), (0, 1), (0, 2)}
$S_3$= {(2,-2), (1,-1), (0, 0), (-1, 1), (-2, 2)}
$S_4$= {(-2, 0), (-1, 0), (0, 0), (1, 0), (2, 0)}.

Then let $S_k^0 = S_k \setminus (0, 0)$ for all k from 1 to 4. In a 5 X 5 window centered at (i, j), for each direction, define $d_{i,j}^{(k)}$ as the sum of all absolute differences of gray level values between $y_{i+s,j+t}$ and $y_{i,j}$ with (s, t) $\in S_k^0$ (k= 1 to 4). Considering that for two pixels whose spatial distance is small, their gray level values should be close; we will weight the absolute differences between the two closest pixels with a large value $\omega_m$, before we calculate the sum. However if $\omega_m$ is very large, it will cause that $d_{i,j}^{(k)}$ is mainly decided by the differences corresponding to $\omega_m$. So let $\omega_m=2$, the reciprocal of distance ratio. Thus we have

$$d_{i,j}^{(k)} = \sum_{(s,t) \in S_k^0} \omega_{s,t} \, |y_{i+s,j+t} - y_{i,j}|, \quad 1 \leq k \leq 4 \quad (1)$$

Where $\omega_{s,t} = \begin{cases} 2 : (s, t) \in \Omega^3 \\ 1 : \text{otherwise} \end{cases}$ (2)

$\Omega^3 = \{(s, t) : -1 \leq s, t \leq 1\}.$ (3)

$d_{i,j}^{(k)}$ is termed as a direction index. Each direction index is sensitive to the edge aligned with a given direction. Then the minimum of these four direction indexes is used for impulse detection, which can be denoted as

$r_{i,j} = \min\{ d_{i,j}^{(k)} : 1 \leq k \leq 4 \}$ (4)

There may be three cases for value of $r_{i,j}$.

1. When the current pixel is a noise free flat region pixel then $r_{i,j}$ is small as because of the four small direction indexes.
2. When the current pixel is an edge pixel then $r_{i,j}$ is small as because at least one of the direction indexes is small.
3. When the current pixel is an impulse pixel then $r_{i,j}$ is large as because of the four large direction indexes.

In definition of $r_{i,j}$, we make full use of the information aligned with four directions. So from the above analysis we can find that by employing a Threshold T, we can Identify the impulse from the noise free pixels, no matter which are in a flat region, edge or thin line. Then we can define the impulse detector as

$y_{i,j}$ is a $\begin{cases} \text{Noisy pixel: if } r_{i,j} > T \\ \text{Noise free pixel: if } r_{i,j} \leq T \end{cases}$ (5)

### III: DWMD Filter

After impulse detection, most median based filters simply replace the noisy pixels by median values in the window. In the proposed technique a new scheme has been introduced based on minimum standard deviation of the four directional pixels.

At first the standard deviation $\sigma_{i,j}^{(k)}$ of gray level values of all $y_{i+s,\,j+t}$ with $(s, t) \in S_k^0$ ($k = 1$ to $4$), is calculated.

Let $l_{i,j} = \underset{k}{\arg\min} \{\sigma_{i,j}^{(k)} : k=1 \text{ to } 4\}$  (6)

where the operator argmin is to find the minimizer of a function. Since the standard deviation describes how tightly all the values are clustered around the mean in the set of pixels, $l_{i,j}$ shows that the four pixels aligned with this direction are the closest to each other. Therefore the center pixel should also be close to them in order to keep the edges (even thin lines) intact in the $l_{i,j}$ direction.

Here a new method has been introduced to make the center pixel (test pixel) as close as possible to the rest four pixels aligned in the direction.

Let the set in the $l_{i,j}$ direction is S= {a, b, c, d, e};

We first replace the middle pixel of the set of pixels by x, to generate the set {a, b, x, d, e}. Now calculate three standard deviations using x= {mean, mean+5 and mean-5}.

Here 5 is used as the set length is 5. The x value is also selected for which the standard deviation of the set is minimum among the 3 standard deviations. It is clear that standard deviation is minimum for x=mean, then we just replace $y_{ij}$ by the mean value and terminates, else we proceed by assigning x by either (mean+5) or (mean-5). Here we iterate the algorithm to minimize the standard deviation by either decreasing x by 5 or by increasing x by 5. The iteration terminates when minimization of standard deviation also stops rather starts increasing.

Let $t_{ij}$ = x.  (7)

Find the element $t_{ij}$ in the initial set S ($l_{i,j}$ directional) and just replace $y_{ij}$ by $t_{ij}$. If not found we assign $t_{ij}$ to its nearest element in the set S, and then $y_{ij}$ is replaced by $t_{ij}$.

This is used to decrease the smoothness and to increase the sharpness of the restored image.

With a single iteration this method of filtering impulse noise from highly corrupted images gives an equivalent result compared to DWM filter and better than its follower paper named a "second order difference based impulse detection method".

### IV: Simulation Results and Discussions

In this section we compare the DWMD filter with a number of existing median-based filters for removal of RVIN.

Figure 2. (a) is the original *Tank* image, 2. (b) and 2. (c) are the 10% and 20% noisy image that of figure 2. (d) and 2. (e) shows restored images from 10% and 20% noise using the proposed technique which shows considerable improvement in both cases.

Again figure 3. (a) shows the original *Map* image and that of figure 3. (b) and 3. (c) are the corrupted image at the level of 10% and 20%. When proposed DWMD is applied on these corrupted images the enhanced image as generated, given in figure 3. (d) and 3. (e) respectively. From these output images it is also clear that the proposed DWMD filter may be useful in removing random valued impulse noise.

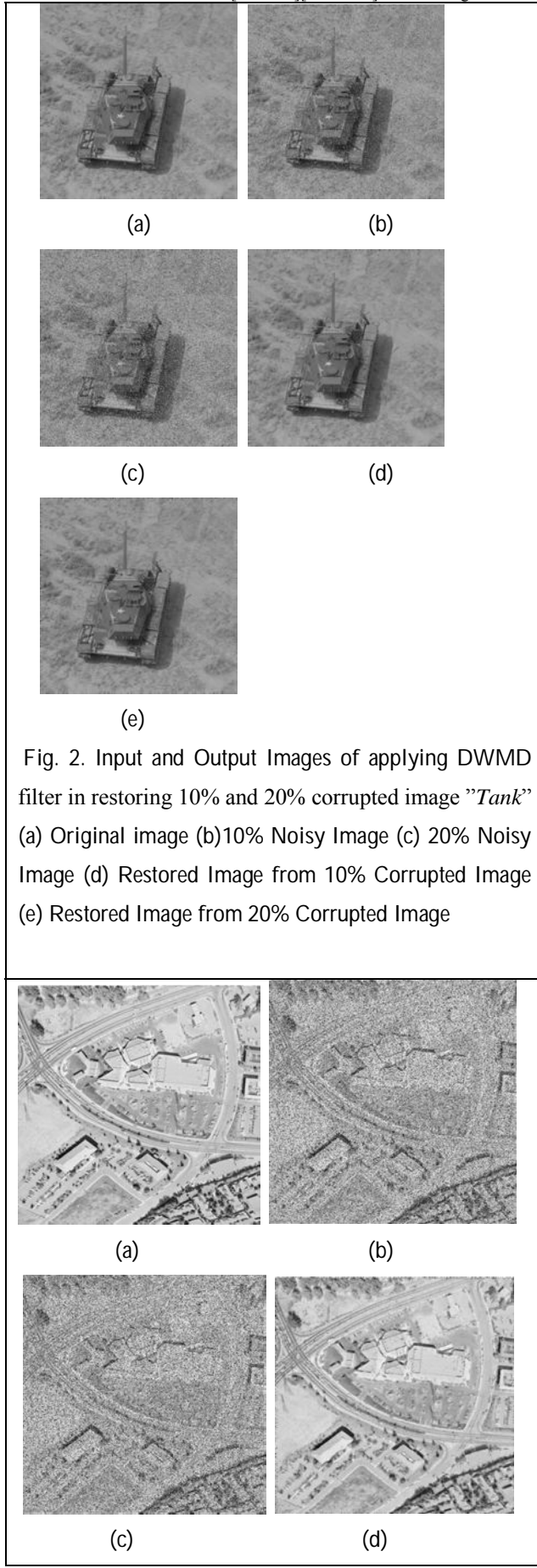

Fig. 2. Input and Output Images of applying DWMD filter in restoring 10% and 20% corrupted image "*Tank*"
(a) Original image (b)10% Noisy Image (c) 20% Noisy Image (d) Restored Image from 10% Corrupted Image (e) Restored Image from 20% Corrupted Image

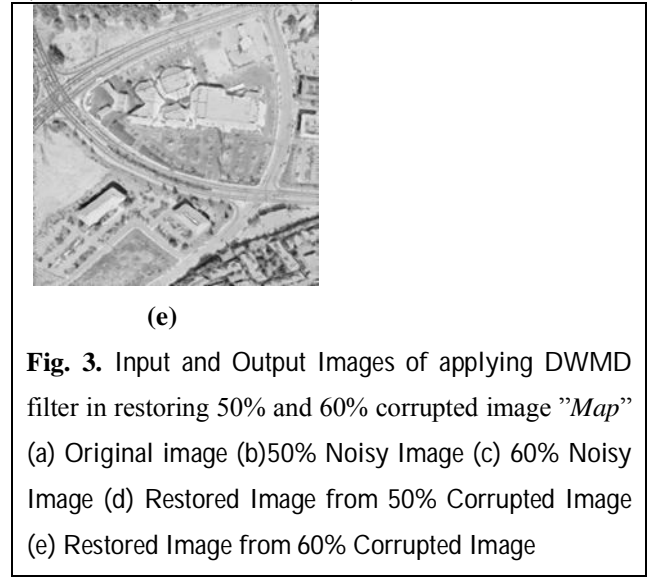

**Fig. 3.** Input and Output Images of applying DWMD filter in restoring 50% and 60% corrupted image "*Map*"
(a) Original image (b)50% Noisy Image (c) 60% Noisy Image (d) Restored Image from 50% Corrupted Image
(e) Restored Image from 60% Corrupted Image

**Table I**
**Restoration Results in PSNR (dB) and Fidelity for Tank and Map Images**

| Noisy Image | PSNR value | Fidelity |
|---|---|---|
| 10% Tank | 34.17 | .998639 |
| 20% Tank | 31.33 | .997379 |
| 50% Map | 23.24 | .990993 |
| 60% Map | 22.55 | .989439 |

**Comparison of image Restoration:** Restoration results are quantitatively measured by Peak Signal to noise Ratio (PSNR) and Image Fidelity (IF). Table I shows the PSNR values of the results obtained by different filters, where 512 X 512 image "*Lena*" corrupted with different noise ratios are used. As we mentioned that our algorithm use single iteration which takes only one uniform threshold value T=256 for all standard images like *Lena, Boat, Bridge* etc. As we can see from Table II, the PSNR (dB) values of restoring 20% to 30% corrupted *Lena* image are not so good, but the proposed filter is outperforms when the image is 40% or more corrupted. PSNR values in Table III for *Bridge* image also shows that the proposed filter works better than any existing filter in restoring 40% or more corrupted image. In Table IV, we have also showed that for 40% or more corrupted *Boat* image the DWMD filter works fine than any existing filter. In Table V we have given the Fidelity Values of Original Image versus Restored Image for *Lena, Boat* and *Bridge* Images for different percent of corrupted image using DWMD

Filter. From this table it can be seen that our filter always preserve its fidelity or faithfulness in restoring corrupted image.

It can be seen that our proposed filter (DWMD) provides the equivalent result to DWM filter or even best result when image is 40% or more corrupted. The DWM filter gives standard result but it does 8 to 10 iteration with decreasing Threshold values to the noisy images. Where as the most recent approach "2nd Order Difference based Filter" has lower computational cost but it does not perform well when noise on the higher side.

Figure 4 and 5 also shows the visual representation of the restoration results using various techniques along with the proposed technique.

Figure 4. (a) shows the original *Lena* image that of 4. (b) is the 60% corrupted image. Five different filters are applied on this *Lena* image and visual effect of the results is compared. Figure 4. (c), 4. (d), 4. (e), 4. (f) are the restored images using SD-ROM, MSM,PWMAD and DWM filter that of figure 4. (g) is the output of the restoration using DWMD filter. Again figure 5. (a) and 5. (b) represent the original and 60% corrupted image that of figure 5. (c), 5. (d), 5. (e) and 5. (f) shows the restored image using ACWMF, PWMAD, DWM and 2nd order difference based filter. Figure 4. (g) shows the restored image using DWMD filter. From all restored results it is clear that proposed DWMD filter obtain better result for high value of impulse noise.

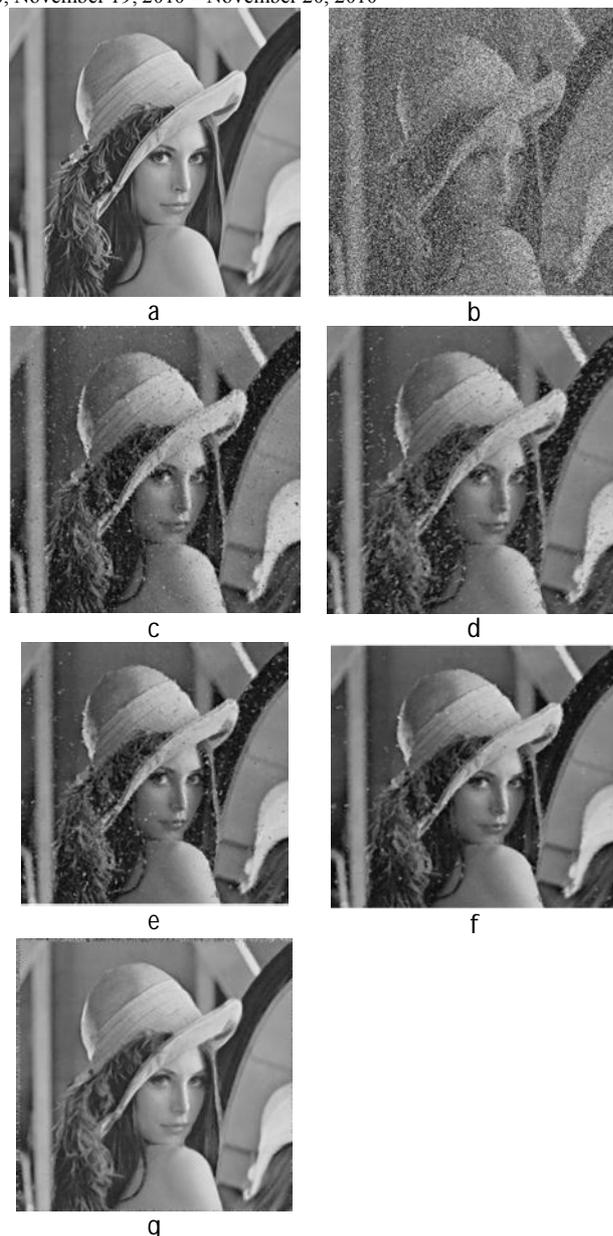

Fig. 4. Results of different filters in restoring 60% corrupted image "*Lena*", (a) Original image (b)Noisy Image (c) SD-ROM Filter (d)MSM Filter (e)PWMAD Filter (f)DWM Filter (g)DWMD Filter

**Table II**

**Comparison of Restoration Results in PSNR (dB) for Lena Image**

| Method | 20% | 30% | 40% | 50% | 60% |
|---|---|---|---|---|---|
| Med Filter[6] | 30.37 | 30.00 | 27.64 | 24.28 | 21.58 |
| SD-ROM[7] | 35.72 | 30.77 | 29.85 | 26.80 | 23.41 |
| PSM Filter[13] | 35.09 | 30.85 | 28.92 | 26.12 | 22.06 |
| MSM Filter[1] | 35.44 | 31.67 | 29.26 | 26.11 | 22.14 |
| ACWM Filter[8] | 36.07 | 32.59 | 28.79 | 25.19 | 21.19 |
| PWMAD Filter[9] | 36.50 | 33.44 | 31.41 | 28.50 | 24.30 |
| Iterative Median Filter[10] | 36.90 | 31.76 | 30.25 | 24.76 | 22.96 |
| **DWM Filter** | **33.81** | **32.43** | **30.64** | **29.14** | **26.57** |
| Second order Difference Based | 34.35 | 32.53 | 30.90 | 28.22 | 24.84 |
| **DWMD Filter** | **33.36** | **32.00** | **31.38** | **29.18** | **26.89** |

**Table III**

**Comparison of Restoration Results in PSNR (dB) for *Bridge* image**

| Method | 40% | 50% | 60% |
|---|---|---|---|
| SD-ROM[7] | 23.80 | 22.42 | 20.66 |
| MSM Filter[1] | 23.55 | 22.03 | 20.07 |
| ACWM Filter[8] | 23.23 | 21.32 | 19.17 |
| PWMAD Filter[9] | 23.83 | 22.20 | 20.83 |
| **DWM Filter** | **24.09** | **23.04** | **21.41** |
| Second order Difference Based Filter | 23.73 | 22.14 | 20.04 |
| **DWMD Filter** | **25.18** | **24.24** | **23.15** |

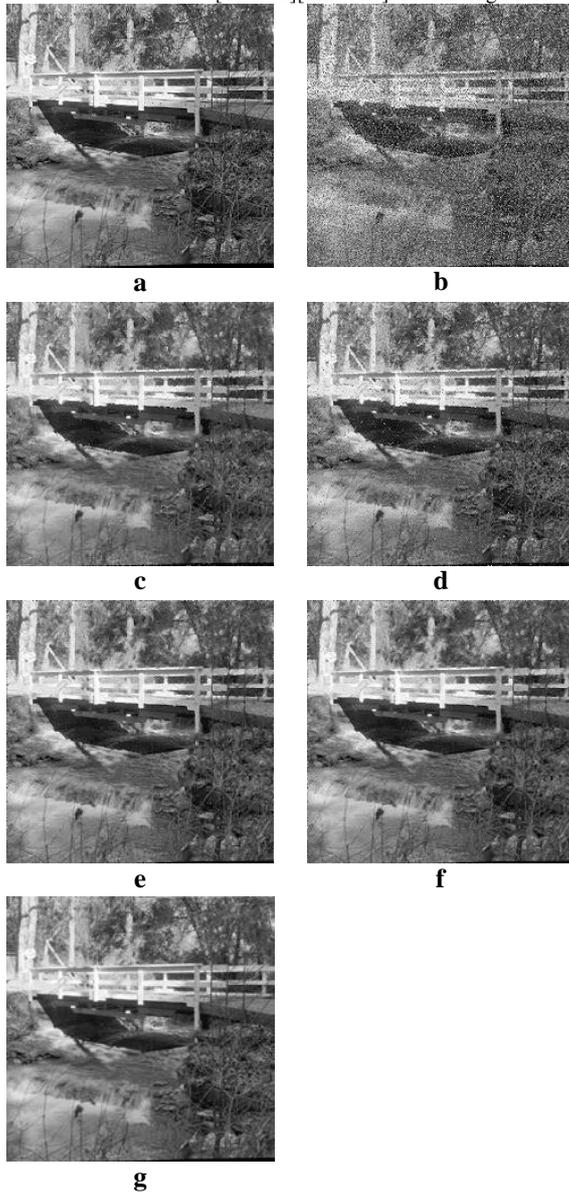

Fig. 5. Results of different filters in restoring 30% corrupted image "Bridge" , (a) Original image (b)Noisy Image (c) ACWMF Filter (d) PWMAD Filter (e) DWM Filter (f) 2$^{nd}$ order Difference Based Filter (g)DWMD Filter

### Table IV

**Comparison of Restoration Results in PSNR (dB) for *Boat* image**

| Method | 40% | 50% | 60% |
|---|---|---|---|
| SD-ROM[7] | 26.45 | 24.83 | 22.59 |
| MSM Filter[1] | 25.56 | 24.27 | 22.21 |
| ACWM Filter[8] | 26.17 | 23.92 | 21.37 |
| PWMAD Filter[9] | 26.56 | 24.85 | 22.32 |
| DWM Filter | 27.03 | 25.75 | 24.01 |
| **DWMD Filter** | **28.15** | **27.08** | **26.59** |

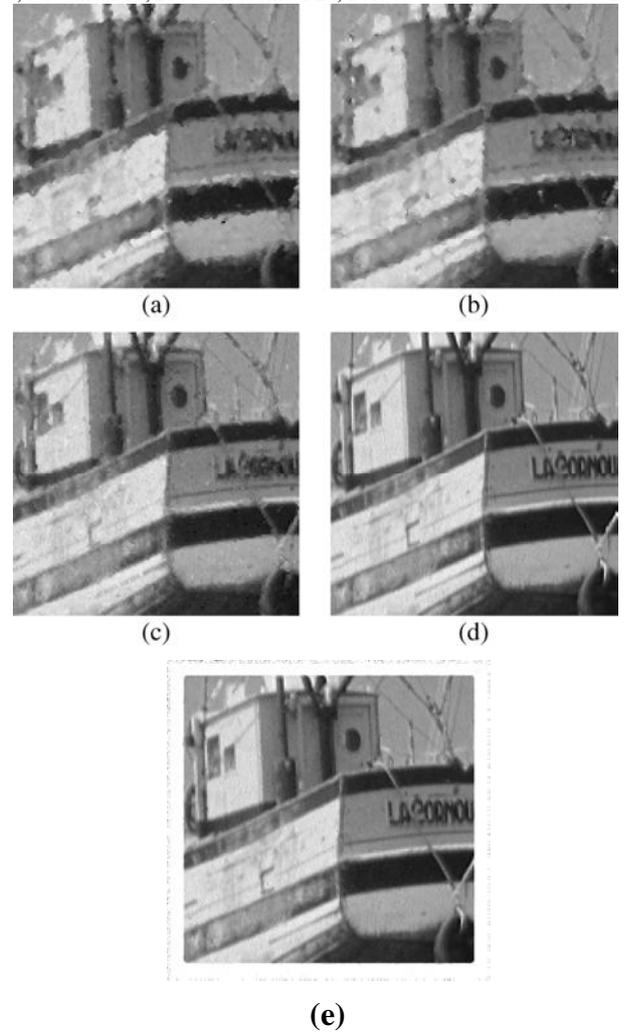

Fig.6. Enlarged areas of different filters in restoring g 40% corrupted image " Boat" (a) SD-ROM(with training) (b) PWMAD Filter (c) DWM Filter (d) Original image (e) DWMD Filter

### Table V

**Fidelity Values of Original Image versus Restored Image for Lena, Boat and Bridge Images for different percent of corrupted image using DWMD Filter**

| Image Name | 30% | 40% | 50% | 60% |
|---|---|---|---|---|
| Lena | .995430 | .992480 | .990956 | .990282 |
| Boat | .991796 | .990228 | .990009 | .989993 |
| Bridge | .989000 | .985061 | .982007 | .979108 |

## V: Conclusion

In this paper we propose a new standard deviation based filter, the DWMD filter, for removing random-valued impulse noise. Simulation results show that the DWMD filter performs better than many existing filters both subjectively and objectively (PSNR), except the DWM, which performs little better when the image is not so noisy. Our algorithm uses the DWM filtering scheme but introduces a new method for noise filtering which little blurs the enhanced image, but it still can preserve the image details such as thin lines. Our letter also provides a table giving the Fidelity (faithfulness) of the Restored image compared to the Original image. From that table we can see that by using DWMD filter we never loose its originality to its restored image.

**Acknowledgements**: The author expresses deep sense of gratitude to the IIPC project of AICTE, of the Dept. of CSE, University of Kalyani, where the computational work has been carried out.